\documentclass[10pt, a4paper]{article}

\usepackage{lrec-coling2024}

\usepackage{times}
\usepackage{latexsym}
\usepackage{graphicx} % DO NOT CHANGE THIS
\usepackage{subfigure}
\usepackage{amssymb,amsthm,dsfont,mathrsfs}
\usepackage{amsmath}
\usepackage{multirow}
\usepackage{mathrsfs}
\usepackage{float}
\usepackage{hyperref}
\usepackage{adjustbox}
\usepackage{booktabs}
\usepackage{multirow}
\usepackage{lscape}
\usepackage{xcolor}
\usepackage{colortbl}
\usepackage{xspace}
\usepackage{tabularx}
\usepackage{verbatim}
\usepackage{url}
\usepackage{bbm}
\usepackage{makecell}
\usepackage{listings}

\usepackage{inconsolata}
\definecolor{Lavender}{rgb}{0.901, 0.901, 0.980}

\title{\vspace*{.5\baselineskip} \normalfont{\vspace*{.5\baselineskip} \textbf{RoCoIns: Enhancing Robustness of Large Language Models through Code-Style Instructions}}}

\name{
      {\normalsize
     Yuansen Zhang$^{\bigstar*}$\thanks{~~Equal contribution.}\quad
     \textbf{Xiao Wang}$^{\bigstar* \dagger}$\quad
     }
    {\normalsize
    \textbf{Zhiheng Xi}$^{\bigstar}$\quad 
    \textbf{Han Xia}$^{\bigstar}$\quad
    } \\
    {\normalsize
    \textbf{Tao Gui}$^{\blacklozenge}$\sthanks{{ }{ }Corresponding Author}\quad
    \textbf{Qi Zhang}$^{\bigstar \dagger}$\quad
    \textbf{Xuanjing Huang}$^{\bigstar}$
    }
}

\address{$^\bigstar$ \normalsize School of Computer Science, Fudan University, Shanghai, China \\
         $^\blacklozenge$ \normalsize Institute of Modern Languages and Linguistics, Fudan University, Shanghai, China \\
           \texttt{\normalsize zhangys22@m.fudan.edu.cn},
           \texttt{\normalsize \{xiao\_wang20,tgui,qz\}@fudan.edu.cn} 
}

\abstract{
Large Language Models (LLMs) have showcased remarkable capabilities in following human instructions.
However, recent studies have raised concerns about the robustness of LLMs when prompted with instructions combining textual adversarial samples.
In this paper, drawing inspiration from recent works that LLMs are sensitive to the design of the instructions, we utilize instructions in code style, which are more structural and less ambiguous, to replace typically natural language instructions.
Through this conversion, we provide LLMs with more precise instructions and strengthen the robustness of LLMs.
Moreover, under few-shot scenarios, we propose a novel method to compose in-context demonstrations using
both clean and adversarial samples (\textit{adversarial context method}) to further boost the robustness of the LLMs. 
Experiments on eight robustness datasets show that our method consistently outperforms prompting LLMs with natural language instructions.
For example, with gpt-3.5-turbo, our method achieves an improvement of 5.68\% in test set accuracy and a reduction of 5.66 points in Attack Success Rate (ASR).
 \\ \newline  }

\begin{document}

\maketitleabstract

\begin{figure}[!htb]
	\centering
	\subfigure[Prompt LLMs with natural language instructions]{
            \label{fig:a}
            \includegraphics[width=0.99\linewidth]{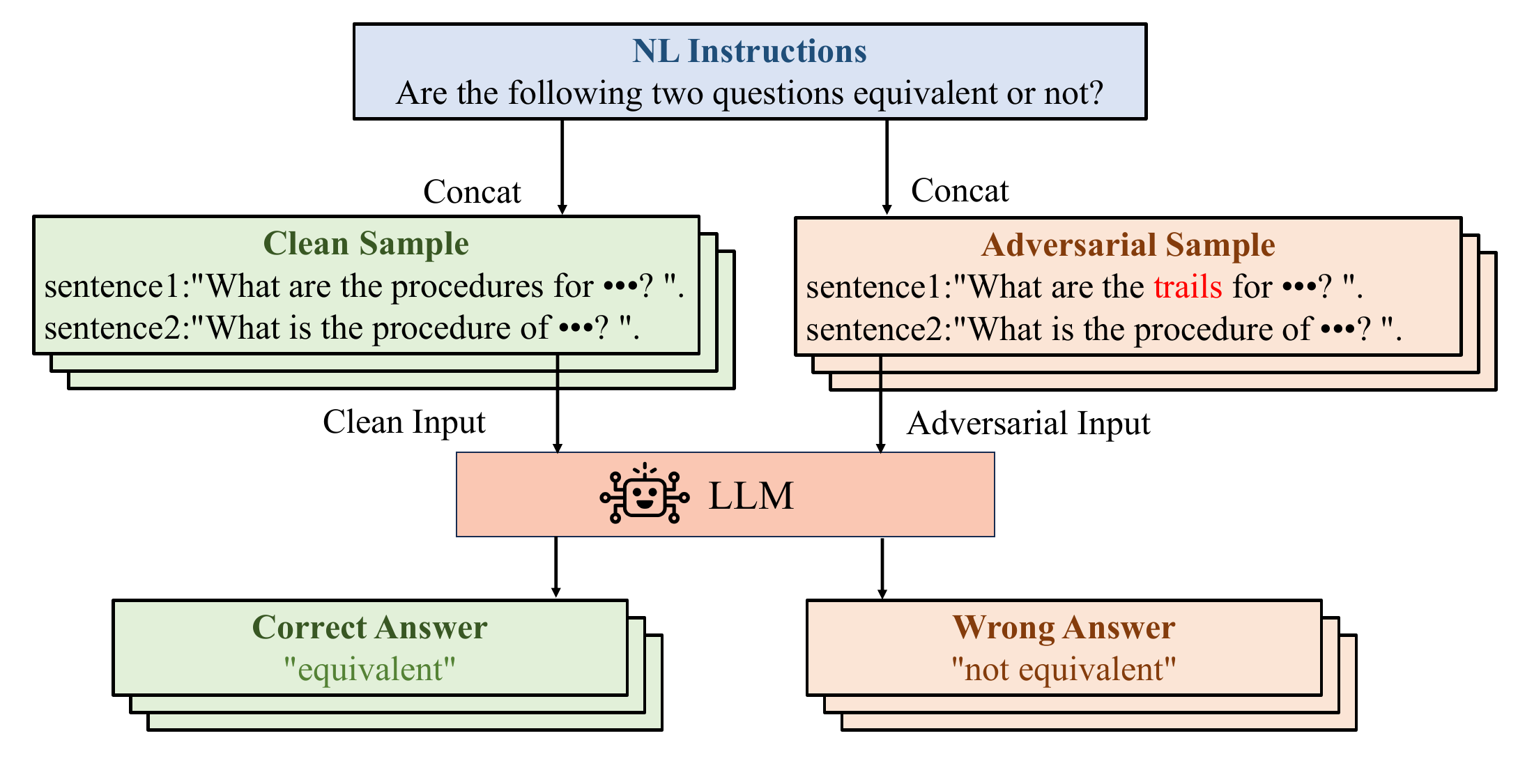}}\\
	\subfigure[Prompt LLMs with code-style instructions]{
            \label{fig:b}
            \includegraphics[width=0.99\linewidth]{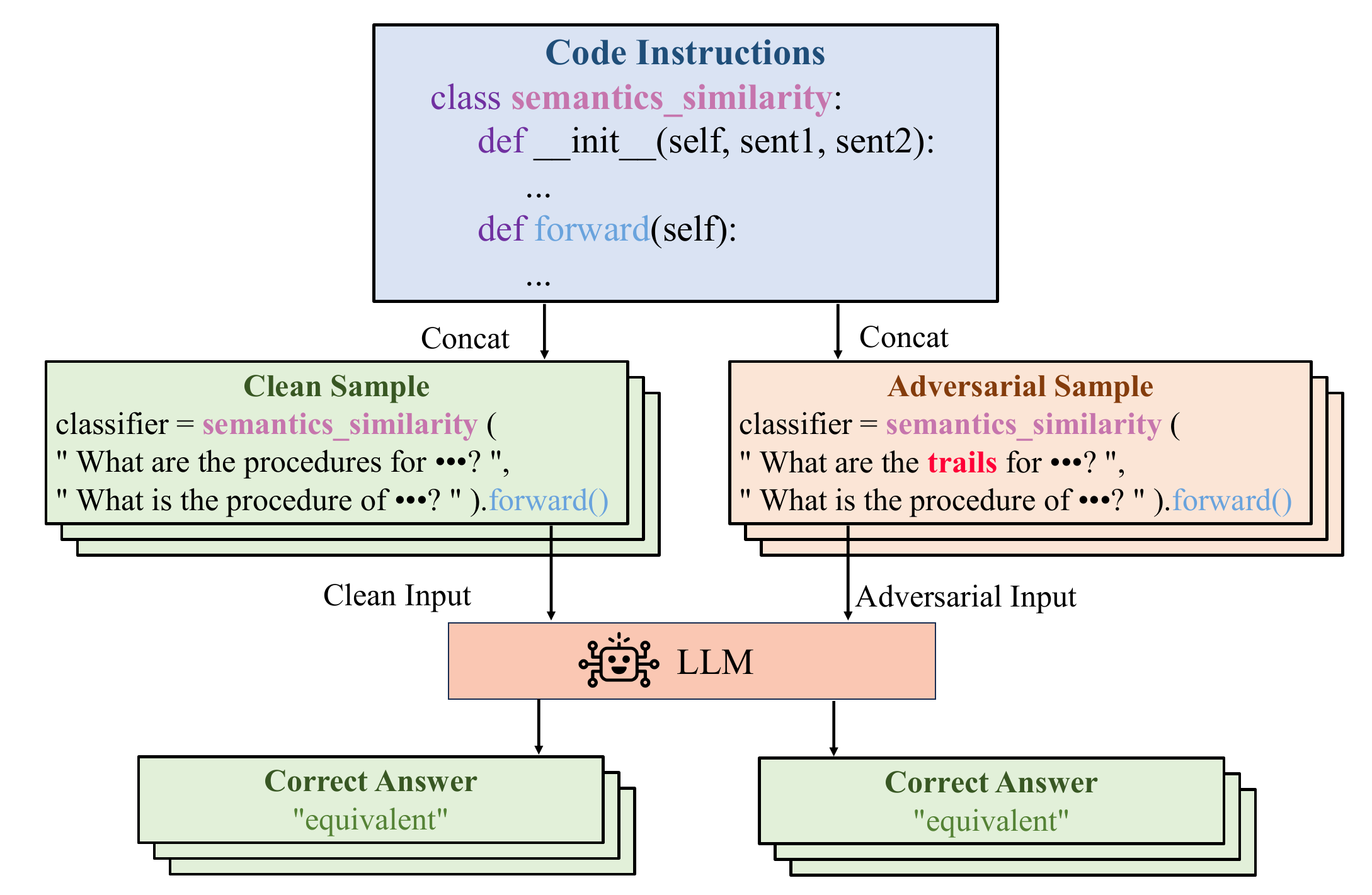}}\\	
	\caption{An illustration of prompting LLMs with natural language instructions and code-style instructions for the \textit{semantics consistent judgment} tasks. The input sample contains a sentence pair.
    We show a clean sample and an adversarial sample, respectively. This code-style instruction can be applied to arbitrary tasks with task-specific design.}
        \label{figure: overview}
\end{figure}

\section{Introduction}

Large language models (LLMs) have shown increasing power in following human instructions and solving various NLP tasks. 
\citep{sanh2022t0, chung2022flant5, ouyang2022InstructGPT, wang2023instructuie, wang2023orthogonal, xi2023rise}.

However, recent evaluations in terms of LLMs have revealed their insufficient robustness when prompted with instructions containing textual adversarial samples, raising concerns about their real-world applications 
\citep{liu2023robustness,wang2023robustness, ye2023comprehensive,chen2023robust}.
By inserting slight perturbations into clean samples at the character, word, or sentence level \citep{gao2018blackbox, ren-etal-2019-generating, Li_Ji_Du_Li_Wang_2019}, the outputs of LLMs occasionally deviate from the expected results. 
For example, in Aspect-based sentiment analysis tasks, when inverting the sentiment polarity of the target aspects, the performance of gpt-3.5-turbo falls by nearly 35\% under zero-shot scenarios \citep{ye2023comprehensive}.

In response to textual adversarial attacks, various adversarial defense methods have been proposed, such as adversarial training \citep{Jiang_He_Chen_Liu_Gao_Zhao_2020}, interval bound propagation \citep{Dvijotham_Gowal_Stanforth_Arandjelovic_O’Donoghue_Uesato_Kohli_2018} and randomized smoothing \citep{Cohen_Rosenfeld_Kolter_2019}.
However, all these methods require parameters update of models, which can be infeasible when it comes to powerful modern LLMs such as GPT-3 \citep{brown2020language} with only APIs provided.
Consequently, limited research has been conducted on enhancing the robustness of such \textit{closed source black-box} LLMs.

To alleviate this problem, we explore enhancing the robustness of LLMs through instructions design.
% We draw inspiration from recent works that the design of instructions can be vitally important to the performance of LLMs \citep{arora2022ask, zhao2021calibrate, gao-etal-2021-making}.
Typically, instructions are formulated using natural language. 
However, the inherent ambiguity of natural language can make LLMs extremely sensitive to instructions, as even slight modifications to the instructions can result in a significant drop in performance \citep{zhao2021calibrate, holtzman2022surface}.
Besides, we believe that introducing adversarial samples into the instructions aggravates this phenomenon and leads to low robustness.
Therefore, it is important to design an instruction format that overcomes these shortcomings.

In this paper, we introduce a novel approach \textbf{RoCoIns}: Enhancing \textbf{Ro}bustness of LLMs through \textbf{Co}de-Style \textbf{Ins}tructions.
The overall framework is shown in Figure \ref{figure: overview}.
%\footnote{In this paper, we mainly focus on attacking samples in the instructions. The detailed difference with other attack formats, such as prompt attacks, can be found in Section ~\ref{Related work:Robustness concerns}.}
We convert the instruction formats from natural language to code style.
The advantages of code, such as being more structural and less ambiguous, provide LLMs with clearer and more concise instructions
 \citep{mishra2023prompting, wang2022code4struct,li2023codeie}, which lead to robustness improvement.
% Besides, LLMs with programs in training corpus are also adept at understanding instructions in code format \citep{chen2021evaluating}.
% Due to these advantages, various tasks have achieved performance improvement by converting the task format from natural language to code, including symbolic and commonsense reasoning, structured prediction, mathematical logic and information extraction \citep{gao2023pal,cheng2023binding,madaan2022language,wang2022code4struct,li2023codeie}.
% These remarkable results motivate us to prompt LLMs with code-type instructions to improve the robustness of LLMs.
% move to relate work
% For example, \citep{wang2022code4struct} exploit the inheritance of classes in Python to allow low-resource event types to acquire extra information from their high-resource siblings. 
% Besides, \citep{li2023codeie} also demonstrates that using prompts of code can be better aligned with the data distribution of models that have been pre-trained with code.
Additionally, we propose the \textit{adversarial context method} to further boost the robustness of LLMs. 
Inspired by \citep{dai2023gpt,vonoswald2022transformers} that in-context learning (ICL) can be considered
as implicit finetuning, we hypothesize that by incorporating both clean and adversarial samples to compose the in-context demonstrations can be viewed as a type of implicit adversarial training. We verify the effectiveness of the method on eight datasets and decrease the average Attack Success Rate (ASR) by 5.66 points with gpt-3.5-turbo.
We conduct further analysis to demonstrate the advantages of using code-style instructions. 

To sum up, our contributions are as follows: 
\begin{itemize}
\setlength{\itemindent}{0em}
\setlength{\itemsep}{0em}
\setlength{\topsep}{-0.5em}
    \item We introduce \textbf{RoCoIns}, a novel approach to enhance the robustness of LLMs against textual adversarial attacks by utilizing code-style instructions.
    % By leveraging the advantages of code, such as structural and less ambiguous, we improve the robustness of LLMs.
    % 措辞
    \item Moreover, we propose the \textit{adversarial context method} to further boost the robustness of LLMs.
    % 第一句拆两句
    \item We conduct experiments on eight robustness datasets and verify the effectiveness of our method, which outperforms prompting LLMs with natural language instructions. 
    % Specifically, with gpt-3.5-turbo on average, our method achieves an improvement of 7\% in robustness test set accuracy and a reduction of 6 points in Attack Success Rate (ASR).
\end{itemize}

\section{Background}

\subsection{Textual Adversarial Attack}
\label{2.1}
Textual adversarial attacks commonly generate explicit adversarial samples by substituting components of sentences with their equivalents while preserving a high degree of semantic similarity \citep{ren-etal-2019-generating,wang2021textflint}.
Given a clean sentence $x = (t_1, t_2, \dots, t_n)$, where $t_i, 1 \leq i \leq n$ denotes each token in the sentence.
$l$ represents its ground truth label.
Textual adversarial attacks replace some original tokens with their counterparts to fool the objective model.
For example, substituting $t_i$ with $\hat{t_i}$ creates an adversary: $\hat{x} = (t_1, t_2, \dots, \hat{t_i}, \dots, t_n)$.
For an adversary, the objective model $F$ generates its label as follows:
\begin{equation}
    \hat{l}=\operatorname{argmax}F\left(\cdot |\hat{x}\right)
\end{equation}
where $\hat{l} \neq l$ means a successful attack.

In this paper, we \textbf{mainly focus on attacking samples rather than instructions}. The detailed difference with other attack formats, such as prompt attacks, can be found in Section ~\ref{Related work:Robustness concerns}.

\subsection{In-context learning with LLMs}
\label{2.2}
Due to the remarkable ICL abilities of LLMs, by providing LLMs with a few demonstration input-output pairs,
they can predict the label for an unseen input without parameter updates.
Formally, we randomly select k sample pairs $\{(x_i,y_i)\}_{i=1}^k$ from the training set 
% and convert them to corresponding task prompt style $\{(x_i^p,y_i^p)\}_{i=1}^k$.
and concatenate them as a string to compose the in-context  demonstrations
$D=x_1 \oplus y_1 \cdot x_2 \oplus y_2 \cdot ... \cdot x_k \oplus y_k$, where $\oplus$ means concatenation between the input and output within a sample and $\cdot$ means concatenation between different samples.
During inference, a new test input $x_{test}$ is appended to the demonstrations, and $D \cdot x_{test}$ is fed into the model for completion and thereby generates an answer $y_{test}^{'}$.

\section{Method}
In this section, we first describe how we recast the instructions  from natural language to code style (Section \ref{3.1}). 
Then we introduce the \textit{adversarial context method} (Section \ref{3.2}).

\begin{figure*}[ht]
    \centering
    % \vspace{-1cm}
    \includegraphics[width=0.99\linewidth]{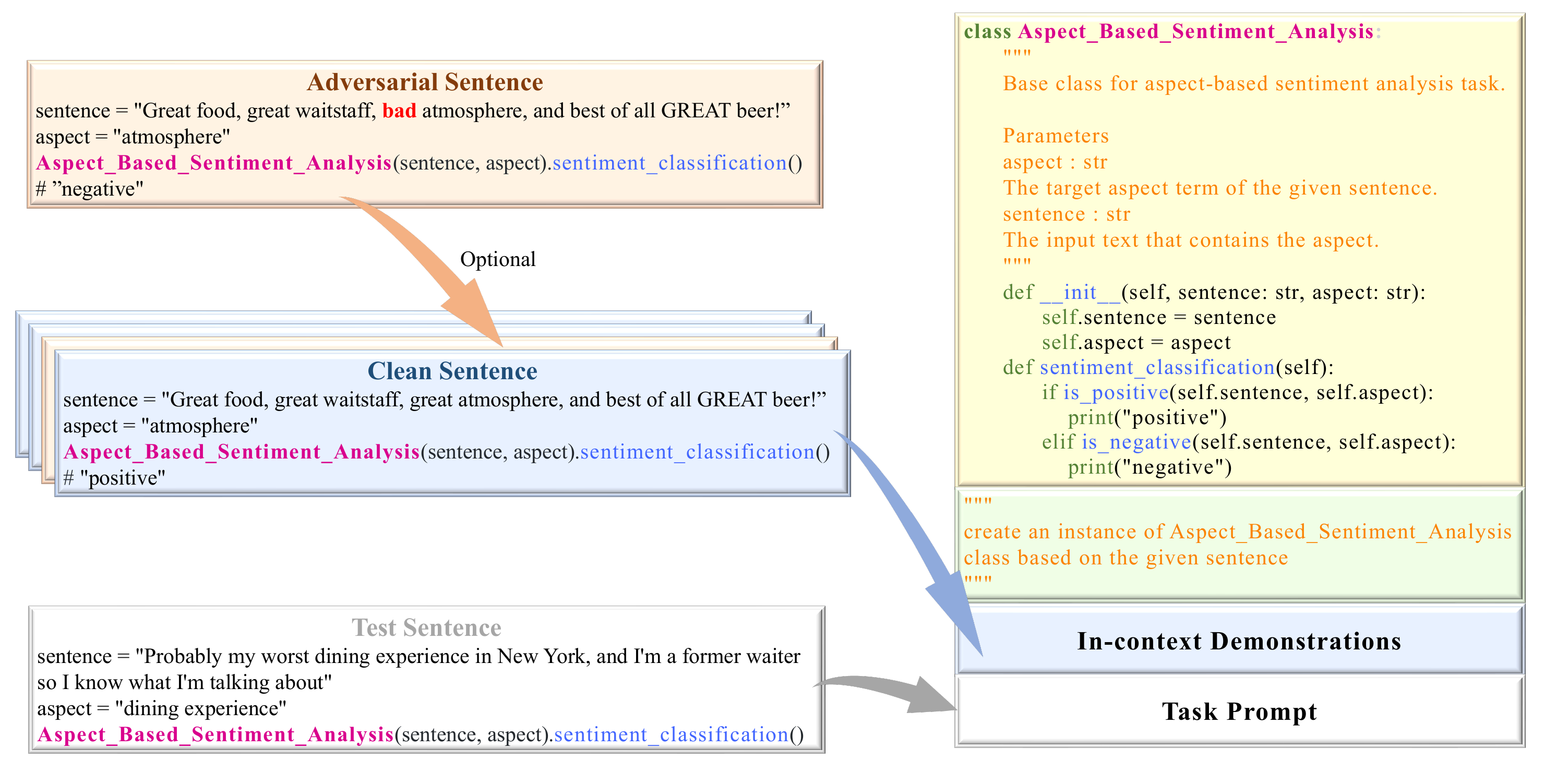}
    \caption{Components of code-style instructions. (\textbf{1}) Class definition mainly contains the class name, annotation, initial function and implementation function. (\textbf{2}) In-context demonstrations consist of k (adversarial) samples in the corresponding code style. (\textbf{3}) Task prompt follows the same format as demonstrations without a ground truth label.}
    \label{fig: example of code prompt}
\end{figure*}
\subsection{Formulating Instructions into Code Style}
\label{3.1}
% We prompt LLMs with program language statements. 
Considering an example in a task with the form $(\mathcal{T},\mathcal{S},\mathcal{L})$, where $\mathcal{T}$ denotes the task instruction, $\mathcal{S}$ refers to input sample and $\mathcal{L}$ represents the corresponding label to be generated. Typically, both $\mathcal{T}$ and $\mathcal{L}$ are expressed in the natural language format. 
However, due to the inherent ambiguity, LLMs have shown extreme sensitivity to these natural language instructions, as even slight modifications to the instructions can lead to a substantial decrease in performance  \citep{zhao2021calibrate, holtzman2022surface}.
In contrast, code-style instructions, which are less ambiguous and more structural, can serve as an alternative to natural language instructions and provide LLMs with more concise instructions.
The primary idea of our method is to convert $\mathcal{T}$  from its original natural language format to a semantically equivalent non-executable pseudo-code format. In this work, we mainly define a \textbf{Python class} to achieve this conversion.
To illustrate our method, we utilize the aspect-based sentiment analysis (ABSA) task as a running example (Figure \ref{fig: example of code prompt}).
ABSA aims to determine the sentiment polarity ("\textit{positive}", "\textit{neutral}" or "\textit{negative}") of an aspect presented in a sentence. 
Our code-style instruction mainly consists of the following components:
% \paragraph{Auxiliary Packages}
% We include the necessary auxiliary packages used in the code.

\paragraph{Class Name} 
First, we convert the explanation of the task into the class name.
The class name can be viewed as a summary of the task.
% Additionally, if functions are employed, we may include all the parameters that are passed into the function.\footnote{In Figure \ref{fig: example of code prompt}, we use Python class to define the task, therefore, parameters are not included here but in the \textbf{init} function. In the experiments section, we explore another code-style instruction by using Python functions format instead of class format to define the  tasks.}

\paragraph{Annotation}
The annotations provide task descriptions that are typically rephrased versions of natural language instructions. Besides, the annotations also contain descriptions of the parameters, including their types and explanations.

\paragraph{Initial Functions}
The initial function defines the input components of this task. For example, in the ABSA task (Figure \ref{fig: example of code prompt}), we define two class instance variables \textit{sentence} and \textit{aspect}, which will be utilized in the subsequent implementation functions.

\paragraph{Implementation Functions}
The implementation functions detail the solution process for the task.
This part is typically constructed based on the annotations and serves as a pseudo-code alternative version of the annotations.
Following \citep{mishra2023prompting}, the implementation function may include sub-task functions, which are usually not explicitly defined and convey their functionalities through descriptive names and parameters.
For example, in the ABSA task (Figure \ref{fig: example of code prompt}), $is\_positive$ and $is\_negative$ are two sub-task functions to determine the sentiment polarity of an aspect word.

\paragraph{Task Prompt}
Once the class is defined, we can utilize it by creating an instance object. 
These objects, accompanied by their properties definitions, compose the task prompt.
Figure \ref{fig: example of code prompt} provides an example of the task prompt.
Typically, we compose an in-context demonstration by concatenating the task prompt with its ground truth label.
Finally, we concatenate the class definition with several in-context demonstrations and a task prompt containing a test sample to construct the model input and expect the model to generate final outputs.

\subsection{Adversarial Context Method}
\label{3.2}
% After that, the model takes $\mathcal{T} \oplus D \oplus x_{test}^p$ as input and generates an answer $y_{test}^p$. 
In this work, we propose the \textit{adversarial context} method to further boost the robustness of LLMs. 
Recent studies have shown that ICL can be regarded as a form of implicit fine-tuning \citep{dai2023gpt,vonoswald2022transformers}. 
\citep{dai2023gpt} theoretically prove that Transformer attention has a dual form of gradient descent and demonstrate that ICL behaves similarly to explicit finetuning from multiple perspectives.
\textbf{Thus, we hypothesize that utilizing both clean and adversarial samples to compose in-context demonstrations can be regarded as a type of implicit adversarial training.}
Formally, following the definitions in Section \ref{2.1} and \ref{2.2}, 
we first transform the clean sample pair $(x_i, y_i)$ and adversarial sample pair $(\hat{x_i}, \hat{y_i})$ into their corresponding code-style format $(x_i^c, y_i^c)$ and $(\hat{x_i^c}, \hat{y_i^c})$.
Then we compose the demonstrations
$D = x_{1}^c\oplus y_1^c \cdot \hat{x_1^c} \oplus \hat{y_1^c} \cdot ... \cdot x_m^c \oplus y_m^c \cdot \hat{x_m^c} \oplus \hat{y_m^c}$ by concatenating both clean and adversarial samples.
We keep the total number of sample pairs unchangeable.

\section{Experiments}
\label{experiments}
\subsection{Experimental setup}
\paragraph{Model}
We conduct experiments mainly using the GPT-3.5 Series models with \texttt{text-davinci-003} and \texttt{gpt-3.5-turbo} from OpenAI \footnote{\url{https://platform.openai.com/docs/models/gpt-3-5}}. 
We choose these two models because GPT-3.5 Series models have shown remarkable code understanding abilities, making them better suited to our proposed method \citep{white2023chatgpt}.
These two models support an input length of up to 8k and 4k tokens, respectively.

\paragraph{Hyperparameters}
We acquire the predictions of the models through OpenAI API \footnote{\url{https://openai.com/api}}.
We prompt LLMs with greedy decode by setting the sampling temperature $t = 0$. 
Besides, we set the max number of generated tokens to 128 tokens.

\paragraph{Datasets}
In this paper, we mainly conduct experiments on two adversarial datasets: \textbf{AdvGLUE} \citeplanguageresource{wang2022advglue} and \textbf{Restaurant} \citeplanguageresource{xing2020tasty}.
AdvGLUE is an adversarial version of the GLUE \citep{GLUE} dataset, consisting of SST-2, QQP, MNLI, QNLI and RTE. We use the test set of AdvGLUE.
Restaurant is an aspect-based sentiment analysis robustness dataset generated from SemEval 2014 Restaurant dataset \citeplanguageresource{pontiki-etal-2014-semeval} by infusing three types of transformations (RevNon, RevTgt and AddDiff) into it \citeplanguageresource{xing2020tasty}. We randomly select 300 samples from the test set of Restaurant for each transformation to compose our test set \textbf{Restaurant-T}.
% The statistic and more details of the datasets can be found in Appendix \ref{sec:Datasets}.

\paragraph{Few-shot Setting}
For each task, we randomly select k samples from the dataset. The choice of k is varied between different tasks according to its number of classes and we explain the reason in Section \ref{5.3}. 
The detailed value of k for each task can be found in Table \ref{table: main experiments}.

\paragraph{Instructions Design}
% All the instructions design can be found in Appendix \ref{Appendix: Prompt}.
The natural language instructions for AdvGLUE are the same with \citep{wang2023robustness} and for Restaurant-T, we choose the same prompts following \cite{chen2023robust}. For code-style instructions, we follow Figure ~\ref{fig: example of code prompt} to construct instructions for different tasks.

\paragraph{Evaluation}
Following \citep{wang2023robustness}, we use Attack Success Rate (ASR) as the evaluation metric for robustness. 
ASR is formally defined as :

\begin{equation}
\label{metrics}
    ASR = \sum_{(x,y) \in T} \frac{\mathbbm{1}[f(A(x))\neq y]}{\mathbbm{1}[f(x)=y]}
\end{equation}
where dataset $T=\{(x_i,y_i)\}_{i=1}^N$ consists of N samples and
$A$ refers to an adversarial attack method, which generates adversarial samples.
In general, the model's robustness against adversarial attacks is inversely proportioned to ASR.
% Besides, we also report the detailed accuracy of each method in Appendix \ref{Appendix B: Detailed Experiments Results}.
All experiments in this paper are conducted 3 times with different demonstrations and we report the mean results.

\begin{table*}[ht]
\centering
\resizebox{\linewidth}{!}{
\begin{tabular}{cc|ccccc|ccc|c|c}
\hline
\multirow{4}{*}{\textbf{Model}} & \multirow{4}{*}{\textbf{Method}} & \multicolumn{8}{c|}{\multirow{2}{*}{\textbf{Dataset(ASR)}}} & \multirow{4}{*}{\textbf{Avg(ASR)}} & \multirow{4}{*}{\textbf{Avg(Acc)}}\\
&& \multicolumn{8}{c|}{} &\\
\cline{3-10}
&& \multicolumn{5}{c|}{\textbf{AdvGLUE}} & \multicolumn{3}{c|}{\textbf{Restaurant-T}} &\\

&& \textbf{SST-2} & \textbf{QQP} & \textbf{MNLI} & \textbf{QNLI} & \textbf{RTE} & \textbf{RevTgt} & \textbf{RevNon} & \textbf{AddDiff} &\\
\hline

\multicolumn{2}{c|}{Random} & $50.0$ & $50.0$ & $66.7$ & $50.0$ & $50.0$ & $66.7$ & $66.7$ & $66.7$ &$58.35$ & $41.67$\\
\hline

\multicolumn{11}{c}{\textbf{Zero-Shot}} \\
\hline
davinci-003 & NL & $44.6$ &$55.1$ &$44.6$ &$38.5$ &$34.6$ &$44.11$	&$20.00$	&$13.19$ &$36.84$ & $-$\\
gpt-3.5-turbo & NL &$39.9$ &$18.0$ &$32.2$ &$34.5$ &$24.74$ & $49.42$ & $36.09$ & $42.67$ & $34.68$ & $-$\\
\hline
\multicolumn{11}{c}{\textbf{Few-Shot}} \\
\hline
\multicolumn{2}{c|}{Shot Number} &$4$&$6$&$6$&$4$&$4$&$6$&$6$&$6$  \\
\hline

\multirow{4}{*}{davinci-003} & NL & $25.39$ & $23.94$ & $25$ & $24.03$ & $15.18$ & $25.09$ & $11.39$ & $10.16$ &$20.02$ &$72.07$\\
& CoT  & $23.07$ & $26.56$ & $23.8$ & $23.62$ & $14.28$ & $24.04$ & $9.35$ & $7.65$ & $19.05$ &$73.43$\\
& \textbf{Code} & $23.97$ & $25.35$ & $22.54$ & $21.73$ & $\textbf{12.65}$ & $23.52$ & $\textbf{5.79}$& $5.08$ & $17.58$ &$75.82$\\
& \textbf{Code+adv} & $\textbf{20.93}$ & $\textbf{22.53}$ & $\textbf{22.33}$ & $\textbf{20.21}$ & $\textbf{12.65}$& $\textbf{21.97}$ & $6.52$ &  $\textbf{4.66}$ & $\textbf{16.47}$ &$\textbf{77.20}$ \\
\hline

\multirow{4}{*}{gpt-3.5-turbo} & NL & $19.23$ & $23.07$ & $21.73$ & $18.75$ & $22.05$ & $29.19$ & $17.03$ & $16.4$ & $20.93$ &$70.45$\\
& CoT & $21.08$ & $20.63$ & $14.85$ & $18.34$ & $\textbf{21.42}$ & $34.67$ & $14.02$ & $14.28$ & $19.91$ &$71.14$\\
& \textbf{Code} & $17.83$ & $18.46$ & $18.55$ & $14.28$ & $21.43$ & $23.35$ & $14.4$ & $\textbf{10.08}$ & $17.29$ &$74.73$\\
& \textbf{Code+adv} & $\textbf{16.43}$ & $\textbf{9.23}$ & $\textbf{14.4}$ & $\textbf{10.71}$ & $22.85$ &$\textbf{22.43}$&$\textbf{13.4}$&$ 12.71$ &$\textbf{15.27}$ &$\textbf{76.13}$\\
\hline
\end{tabular}
}
\caption{Experiments performances on AdvGLUE and Restaurant-T datasets. We report the $ASR (\downarrow)$ for each method. We also report the average accuracy($Avg(Acc)\uparrow$) in the last column. In this table, Our methods and the best results are highlighted in \textbf{bold}. NL and \textbf{Code} refer to prompting with natural language and code-style instructions, respectively. \textbf{Code+adv} refers to our proposed \textit{adversarial context method}.}
\label{table: main experiments}
\end{table*}

\begin{table}[ht]
    \centering
    \resizebox{\linewidth}{!}{
    \begin{tabular}{cc|c|c}
    \hline
    Model & Method & Clean & Adversarial \\
    \hline
    \multirow{3}{*}{davinci-003} & NL & $86.78$ &$72.07$\\
    & \textbf{Code} & $87.25(+0.47)$ & $75.82(+3.75)$ \\
    & \textbf{Code+adv} & - & $77.20(+5.13)$ \\
    
    \hline

    \multirow{3}{*}{gpt-3.5-turbo} & NL & $85.61$ & $70.45$\\
    & \textbf{Code} & $86.13(+0.52)$ &$74.73(+4.28)$\\
    & \textbf{Code+adv} & - & $76.13(+5.68)$ \\
    
    \hline

    \end{tabular}
    }
    \caption{Average Accuracy on the 8 clean and adversarial datasets for NL, \textbf{Code} and \textbf{Code+Adv} methods.}
    \label{table:Avg Acc}
\end{table}

\paragraph{Baselines} 
\begin{itemize}
\setlength{\itemsep}{2pt}
\setlength{\parsep}{0pt}
\setlength{\parskip}{0pt}
    \item [1)] \textbf{Zero-shot NL Prompting} To evaluate the impact of few-shot prompting on enhancing the robustness of language models (LLMs), we consider zero-shot natural language prompting as a baseline for comparison. The zero-shot results of AdvGLUE are from \citep{wang2023robustness}.

    \item [2)] \textbf{Few-shot NL Prompting} Under few-shot settings, we compare our approach with natural language prompting. By using the same natural language instructions with zero-shot prompting, we additionally provide LLMs with a few [Problem, Answer] samples to help LLMs better understand the tasks and standardize output formats.
    \item [3)] \textbf{Few-shot CoT Prompting} Since Chain-of-Thought (CoT) \citep{wei2023chainofthought} has verified its effectiveness in improving performance on various tasks, we also incorporate CoT as a baseline to explore its effectiveness in robustness improvement. Specifically, we provide LLMs with a set of [Problem, Rational, Answer] samples to encourage LLMs to think step-by-step and generate final answers.
\end{itemize}

\subsection{Results}
\label{4.2}
% In this section, we present the main experiments on AdvGLUE and Restaurant-T datasets. We report the ASR of each method in Table \ref{table: main experiments}. Additionally, a detailed report of the accuracy for all datasets before and after adversarial transformation can be found in Appendix \ref{Appendix B: Detailed Experiments Results}.

\paragraph{NL instructions vs. Code-style instructions}
% We then compare the performance of using natural language instructions and code-style instructions to prompt LLMs. 
As shown in Table \ref{table: main experiments}, prompting LLMs with code-style instructions consistently outperforms prompting with natural language instructions. 
Specifically, code-style instructions result in a 2.44 and 3.64 point reduction in ASR on \texttt{text-davinci-003} and \texttt{gpt-3.5-turbo}, respectively.
We also provide the average accuracy in Table ~\ref{table:Avg Acc}.
We observe a slight improvement by using code-style instructions when prompting LLMs with clean samples($0.47$ and $0.52$), but a relatively huge improvement with adversarial samples($3.75$ and $4.28$), \textbf{which indicates the advantages of using code-style prompts when faced with adversarial samples}.
A more detailed analysis of the advantages of code-style instructions is provided in Section \ref{Analysis}.

\paragraph{Adversarial context further enhances the robustness}
We further demonstrate the effectiveness of our proposed \textit{adversarial context method}. 
% By incorporating adversarial samples into the in-context demonstrations, we further boost the robustness of LLMs. 
Compared to natural language prompting, the adversarial context method leads to a decrease of 3.55 points in ASR for \texttt{text-davinci-003} and 5.66 points for \texttt{gpt-3.5-turbo}.
Besides, from Tabel ~\ref{table:Avg Acc}, incorporating our adversarial context method results in a significant improvement in accuracy.
Specifically, there is an improvement of 5.13 points with \texttt{text-davinci-003} and 5.68 points with \texttt{gpt-3.5-turbo}.
We hypothesize that the improvement brought by adversarial samples could be attributed to the implicit adversarial training through in-context learning.
Additionally, introducing adversarial samples prompts the LLMs to recognize specific adversarial attacks, such as spelling errors and word substitutions. The findings also suggest that more advanced models, like \texttt{gpt-3.5-turbo}, potentially benefit more from code-style instructions and the adversarial context method than \texttt{text-davinci-003}.

\paragraph{Zero-Shot vs. Few-shot}
As shown in Table \ref{table: main experiments}, zero-shot prompting exhibits low robustness on both \texttt{text-davinci-003} and \texttt{gpt-3.5-turbo}. In particular, for some tasks, the LLMs perform only slightly better or even worse than random guessing (for example, QQP on \texttt{text-davinci-003}).
However, when prompting LLMs with additional in-context demonstrations, the robustness of LLMs improves by a large margin. By few-shot prompting, the average ASR of \texttt{text-davinci-003} and \texttt{gpt-3.5-turbo} decrease by 16.82 and 13.75 points, respectively.
This indicates the strong few-shot learning abilities of LLMs. \textbf{By leveraging only a few examples, LLMs can better understand the task and yield stronger robustness}.  

\begin{figure}[ht]
    \centering
    \includegraphics[width=0.99\linewidth]{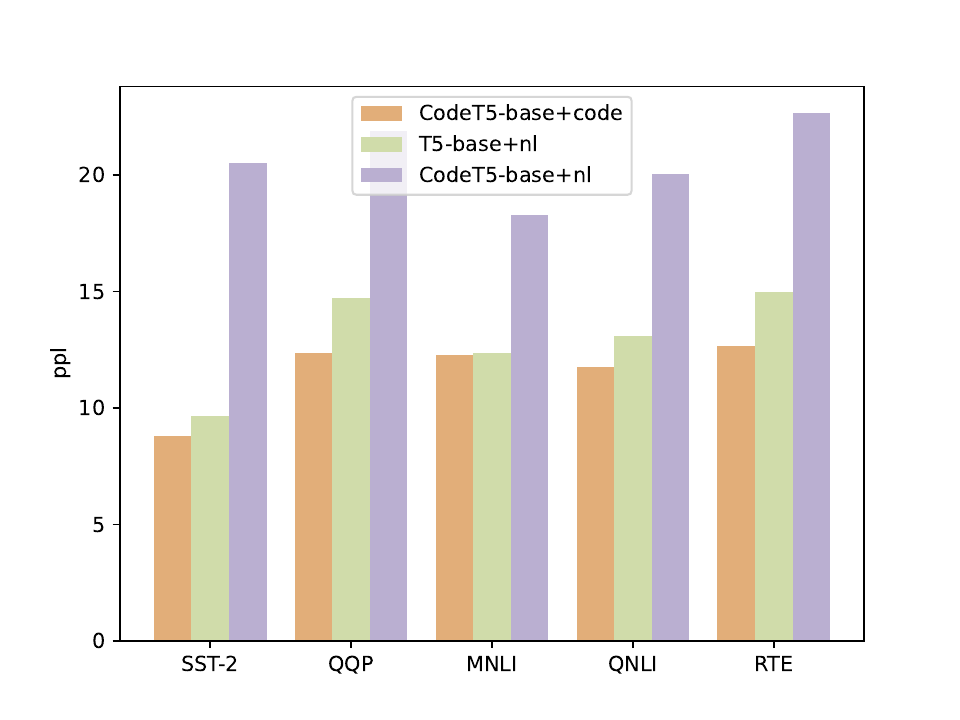}
    \vspace{-1cm}
    \caption{Perplexity for AdvGLUE dataset on T5-base with natural language instructions and CodeT5-base with both natural language and code-style instructions. We report the logarithm of their initial values.}
    \label{fig:ppl}
\end{figure}
\paragraph{Chain-of-thought helps with robustness}
We then explore whether CoT can help improve the robustness of LLMs. \textbf{By promoting LLMs to think step-by-step to generate final answers, we find that for most tasks, LLMs showcase better robustness than directly prompting LLMs to generate the final answer}. 
On average, we present a decline in ASR by 0.97 and 1.02 points with \texttt{text-davinci-003} and \texttt{gpt-3.5-turbo}, respectively.
However, we also observe a decrease in robustness on specific datasets, such as RevTgt on \texttt{gpt-3.5-turbo}. By analyzing the reasoning steps, we find that the over-complicated and neutral-oriented reasoning process contributes to the failure of CoT.

\section{Analysis}
\label{Analysis}

\subsection{Perplexity: Code vs NL}
To take a closer look at the advantages of using code-style instructions, we hypothesize that utilizing code-style instructions can provide LLMs pre-trained on code data with more precise instructions, consequently resulting in performance improvement. 
To verify our hypothesis, we compare the perplexity of a pre-trained language model on the natural language instructions and a pre-trained code model on both the natural language and code-style instructions.
Specifically, we calculate the mean perplexity \textit{ppl} of a dataset T consisting of N samples using the following formula:
\begin{equation}
    ppl_M(T) = \frac{1}{n}\sum_{(x,y)\in T} \prod_{i=1}^m P_M(y_i|y_1\dots y_{i-1},x)^{-\frac{1}{m}}
\end{equation}
where m refers to the length of the generated tokens.
For each sample $(x,y)$ in T, we convert it to both natural language format $(x_{nl},y_{nl})$ and code-style format $(x_c,y_c)$ and then calculate the perplexity with two models $M_{nl}$ and $M_{c}$.
A lower perplexity suggests the models are less confused by the instructions and output format.

\begin{figure}[ht]
    \centering
    \includegraphics[width=0.99\linewidth]{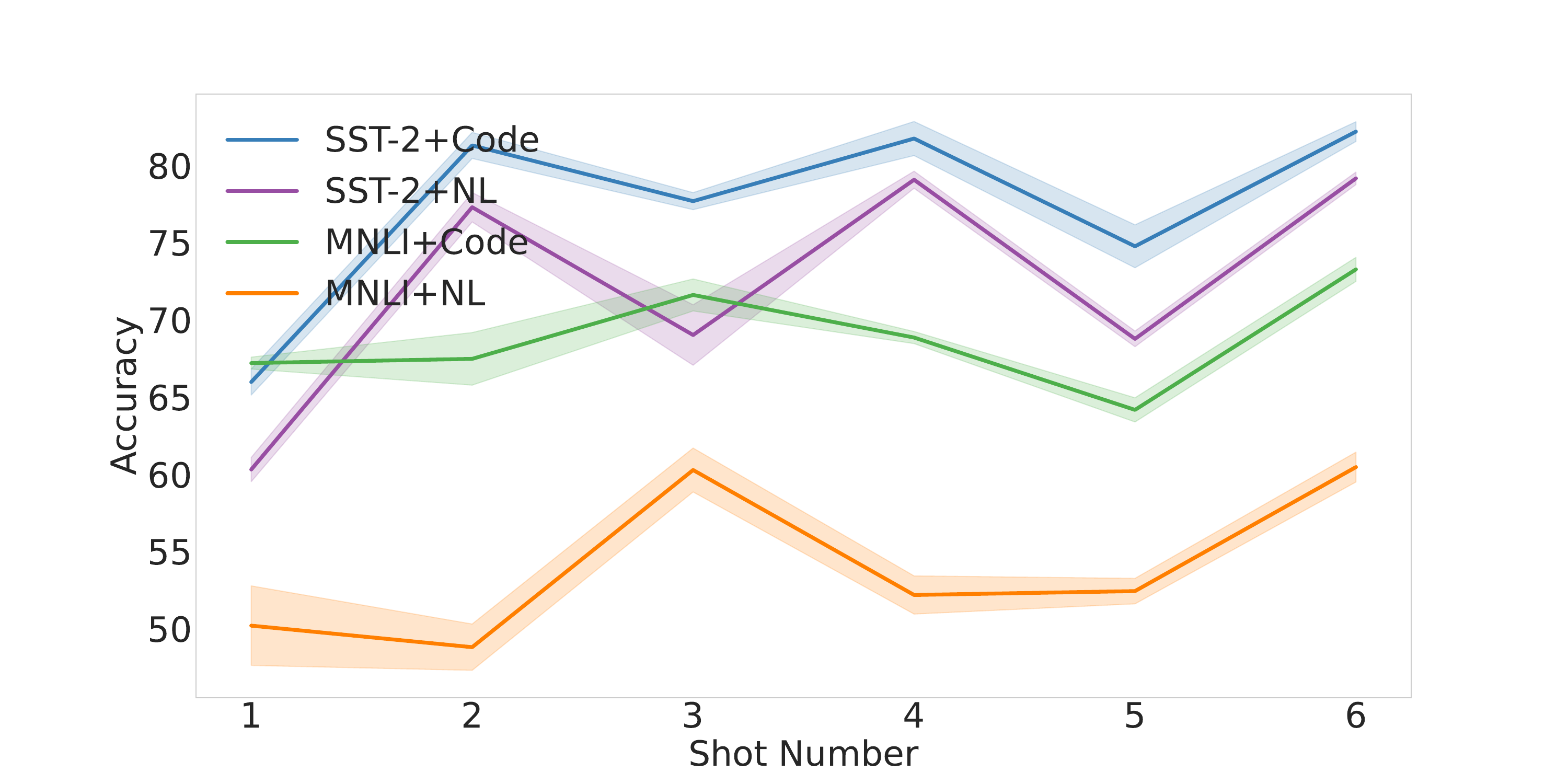}
    \vspace{-0.5cm}
    \caption{Accuracy with the different number of in-context demonstrations on SST-2 and MNLI adversarial dataset. The experiment is conducted on gpt-3.5-turbo.}
    \label{fig:shot_number}
\end{figure}

Due to the \textit{black-box} features of LLMs, obtaining logits directly from LLMs is challenging.
Therefore, following \citep{li2023codeie}, we use T5 \citep{T5} and CodeT5 \citep{CodeT5}, which are further pre-trained on code data for T5, to compute perplexities.
We use the AdvGLUE dataset to calculate the perplexity of T5-base with natural language instructions and CodeT5-base with both natural language and code-style instructions.
As shown in Figure \ref{fig:ppl}, utilizing the pre-trained code model with code-style instructions consistently results in the lowest perplexity, surpassing the performance of using natural language instructions in both the pre-trained language and code models.
\textbf{This observation suggests that converting instructions into code style better align with the pretraining data distribution for pre-trained code models}.

\
\begin{figure*}
    \centering
    \includegraphics[width=0.99\linewidth]{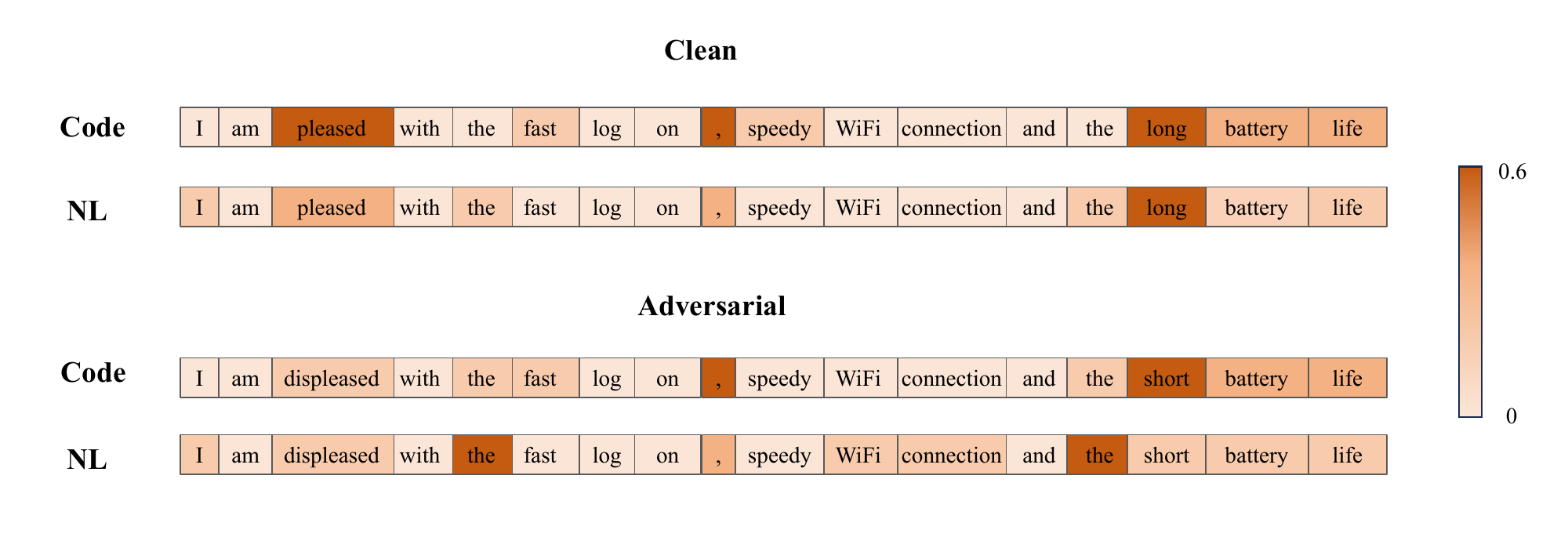}
    \vspace{-0.8cm}
    \caption{Visualization of a sample's gradient on each word when fine-tuning CodeT5 with code-style instruction and T5 with natural language instruction, respectively. The sample is selected from the Restaurant-T dataset with both its clean and adversarial versions. The sample aims to determine the sentiment polarity of the aspect "\textit{battery life}" in the sentence with "\textit{positive}" or "\textit{negative}".}
    \label{fig: visualization}
\end{figure*}

\begin{table}[ht]
    \centering
    \begin{tabular}{cccc}
    \hline
    \textbf{Prompt} & \textbf{SST-2}& \textbf{MNLI} & \textbf{QNLI} \\
    \hline
    NL & $19.23$&$ 21.73$& $18.75$\\
    NL(complicated) &$19.45$& $21.64$ & $18.97$\\
    \hline
     \cellcolor{Lavender}Class Exec & \cellcolor{Lavender}$\textbf{17.83}$ & \cellcolor{Lavender}$18.55$ & \cellcolor{Lavender}$14.28$ \\
    Class Init &$20.93$ &$\textbf{16.12}$& $16.52$\\
    Func Exec & $18.6$ & $17.52$ & $\textbf{14.03}$\\
    \hline
         
    \end{tabular}
    \caption{ASR for different code-style instructions design. "class exec" is the code format used in our main experiments and is highlighted in light grey. The experiment is conducted on gpt-3.5-turbo.}
    \label{table: instruction format}
\end{table}

\subsection{Ablation Studies}
\label{5.2}
\subsubsection{Different Code-style Instructions}
To explore whether using code-style instructions can generally obtain better robustness, following \citep{wang2022code4struct} and \citep{li2023codeie}, we design two different code-style instructions \textit{class init} and \textit{func exec}. The \textit{class init} provides LLMs with incomplete code with partial parameter input of the class as triggers to prompt LLMs to complete the code. The \textit{func exec} converts the class definition into function definition. 
% All instructions designs can be found in Appendix \ref{Appecdix B.3:Different code-style instructions}. 
We also design a more complicated NL prompt for comparison.
We report the results in Table \ref{table: instruction format}, from which we find that code-style instructions almost always outperform natural language instructions with lower ASR, showcasing the overall superiority of code-style instructions. Besides, using more complicated natural language instructions does not help with the robustness of LLMs.

\subsubsection{Number of In-context Demonstrations}
\label{5.3}
To investigate the influence of different numbers of demonstrations, we perform experiments on SST-2 and MNLI, ranging from 1 to 6 shots. As shown in Figure \ref{fig:shot_number}, we find that the completeness and balance of labels are significant for the task performance. For example, for SST-2 with two labels \textit{positive} and \textit{negative}, when prompting with an even number of demonstrations (\textit{positive} and \textit{negative} are both included and the number of each label is equal), we consistently get better results than those with incomplete labels (1-shot) or with imbalanced labels (3-shot, 5-shot).
Therefore in Section \ref{4.2}, we choose shot number k according to the number of labels for different tasks.

\subsubsection{Different Part of Code Instructions}
To assess the influence of different parts of our code-style prompts, we conduct three transformations on the instructions. 
As shown in Table ~\ref{table:Different part}, randomly replacing the \textit{Class Name} and \textit{Sub-task Name} has minimal impact on performance while removing the annotation leads to a slight decline in performance. This shows the toughness of our code-style instructions against different disturbance.

\begin{table}[h]
    \centering
    \begin{tabular}{lcc}
    \hline
    \textbf{Component} & \textbf{SST-2} & \textbf{RTE}  \\
    \hline
    Code Instructions & $83.1$ & $74.07$ \\
    - Class Name & $82.43(-0.67)$ & $72.83(-1.24)$ \\
    - Sub-task Name & $81.08(-2.02)$ & $75.31(+1.24)$\\
    - Annotation & $78.38(-4.72)$ & $69.14(-4.93)$ \\

    \hline
    \end{tabular}
    \caption{Results for different parts of code-style instructions. We report the Acc with gpt-3.5-torbo. We conduct three separate experiments: randomly replacing the \textit{Class Name}, randomly replacing the \textit{Sub-task Name} and removing the \textit{Annotation}.}
    % \vspace{-0.7cm}
    \label{table:Different part}
\end{table}

\subsection{Visualization Analysis}
To further investigate which part the model focuses on, we select a sample from the Restaurant-T dataset with both its original and adversarial forms. 
We then utilize natural language instructions and code-style instructions to wrap the sample and then fine-tuning them using T5-base and CodeT5-base, respectively. 
We extract the gradients of each token from the model embedding layer and average them across dimensions.
Moreover, we normalize the gradients within the sentence and obtain final gradients.
We visualize the gradients in Figure \ref{fig: visualization}.
Since the gradients can reflect how much the model focuses on the token \citep{li2016visualizing, Madsen_2022},
for clean sentence, both natural language instruction and code-style instruction focus on the right word "long".
While for adversarial sentence, code-style instruction with CodeT5-base can still lead the model to pay attention to the phrase "\textit{short battery life}". However, using natural language with T5-base,  the model focuses on irrelevant phrases such as "\textit{the}".
Therefore, code-style prompts may help the model focus more on the important part of a sentence.

\subsection{Discussion for User-friendliness}
Although showing impressive performance with code-style instructions, it may be difficult for non-professional users to transform the prompts into code. Actually, the code-style prompts we design are straightforward and can be easily adapted to arbitrary tasks with similar structures. Users can follow our structure either manually or through LLM-based methods (e.g., utilizing LLMs) to construct task-specific code-style prompts. 
To verify the simplicity of our method, we select several tasks (SST-2, MNLI, RTE) and concatenate them to prompt GPT-4 to generate code-style prompts for new tasks. The results are shown in Table ~\ref{table:user-friendliness}. We observe that LLM-based prompts almost match the performance of manually designed prompts.

\begin{table}[h]
    \centering
    \begin{tabular}{ccc}
    \hline
    \textbf{New Task} & \textbf{LLM-based} & \textbf{Manual}  \\
    \hline
    QNLI & $72.07_{\pm 2.1}$ & $73.64$ \\
    RevTgt & $70.22_{\pm 2.7}$ & $71.33$ \\
    \hline
    \end{tabular}
    \caption{Results for LLM(GPT-4)-based prompts with gpt-3.5-turbo. We report the Acc of new tasks QNLI and RevTgt. "LLM-based" refers to GPT-4 generated prompts. "Manual" refers to our manually designed prompts.}
    \label{table:user-friendliness}
\end{table}

\section{Related Work}
\paragraph{Textual Adversarial Attacks/Defenses}
Textual attacks typically generate explicit adversarial examples by adding small perturbations into clean examples while maintaining lexical correctness, grammatical correctness and semantic similarity \citep{Generating,wang2021textflint}. These adversarial methods can be divided into character-level \citep{gao2018blackbox}, word-level \citep{ren-etal-2019-generating} and sentence-level \citep{Li_Ji_Du_Li_Wang_2019}.
In response to adversarial attacks, various defense methods have been proposed\citep{Jiang_He_Chen_Liu_Gao_Zhao_2020,wang2022miner}. Adversarial training \citep{FreeLB} is a widely adopted approach that iteratively solves a two-layered min-max optimization problem.
Interval bound propagation \citep{Dvijotham_Gowal_Stanforth_Arandjelovic_O’Donoghue_Uesato_Kohli_2018} is proposed to find worst-case adversaries.
Besides, randomized smoothing \citep{Cohen_Rosenfeld_Kolter_2019} and adversarial detection \citep{alshemali2019toward, mozes2021frequencyguided} are also popular methods in defending adversarial attacks. However, all these methods require parameter updates and can be unattainable when faced with \textit{closed source black-box} LLMs. Therefore, in this work, we propose a novel approach to enhance the robustness of LLMs through instructions design without the need for parameter updates.

\paragraph{Robustness Concerns for LLMs}
\label{Related work:Robustness concerns}
While the progress of LLMs has shown remarkable abilities in following human instructions and generating safe content, recent works pose concerns about the robustness of LLMs \citep{liu2023robustness, shi2024navigating}\citeplanguageresource{wang2022advglue}. 
Attacks based on prompts have showcased the possibility of attacking LLMs with adversarial prompts \citep{zhu2023promptbench,ni2023evaluating}. For example, "jailbreak" attempts to modify clean prompts to elicit undesirable responses from LLMs \citep{wei2023jailbroken}.
\citep{zou2023universal} find adversarial attacks on aligned language models and prove their universal and transferable attack ability. 
\textbf{In this paper, our attack aims at destroying the samples while keeping the prompts clean, which is different from prompts attacks that focus on destroying prompts}.
% and backdoor attacks \citep{gu2019badnets,hintersdorf2023balancing}.

\paragraph{In-context Learning}
With the model scale growing, directly fine-tuning the model can be extremely expensive due to storage and time complexities \citep{rae2022scaling,chowdhery2022palm,smith2022using}. 
Alternatively, in-context learning (ICL) has been verified to be an effective way for LLMs to learn a new task by conditioning on a few training examples \citep{brown2020language}.
There are already lots of works demonstrating the perfect ICL abilities of LLMs in solving complex tasks, such as solving mathematical reasoning problems \citep{wei2023chainofthought}.
On the other hand, plenty of studies have investigated the mechanism behind ICL. 
\citep{xie2022explanation} explained ICL from the perspective of implicit Bayesian inference.
\citep{dai2023gpt,vonoswald2022transformers} viewed ICL as implicit fine-tuning and theoretically demonstrated that Transformer attention has a dual form of gradient descent.
% \citep{li2023transformers} abstracted ICL as an algorithm learning problem and showed that Transformers can implement a proper function class through implicit empirical risk minimization for the demonstrations.
Therefore, in this work, we hypothesize that incorporating adversarial samples as demonstrations can be viewed as a form of adversarial training and propose the \textit{adversarial context method}.

\paragraph{Code-style Instructions for Different Tasks}
Due to the advantages of code format, plenty of works have used code-style instructions to tackle complex tasks. \citep{gao2023pal} used programs to split the decomposition and computation of a mathematical problem.
\citep{wang2022code4struct} leverage LLMs text-to-structure translation capability to solve structured prediction tasks. \citep{madaan2022language} frame structured commonsense reasoning tasks as code generation tasks. \citep{li2023codeie} recast the structured output of IE tasks in the form of code instead of natural language. Besides, similar to our work, \citep{mishra2023prompting} utilizes pseudo-code instructions to prompt pre-trained models such as CodeGen \citep{nijkamp2023codegen} to improve the performance of pre-trained language models.

\section{Conclusion}
In this paper, we propose \textbf{RoCoIns} to utilize code-style instructions instead of natural language instructions to enhance the robustness of \textit{closed source black-box} models against textual adversarial attacks. 
% Due to the inherent ambiguity of natural language, LLMs can exhibit a high sensitivity and low robustness when confronted with natural language instructions with adversarial attacks. 
Instructions in code style, which are more structural and less ambiguous than natural language instructions, provide LLMs with more precise instructions. 
Besides, we propose \textit{adversarial context method} to further boost the robustness. Experiments show that our method consistently outperforms prompting LLMs with natural language instructions under the few-shot setting. We conduct further analysis to verify the advantages of using code-style instructions.

\section{Limitations}
Due to the limitation of \textit{closed source black-box} models, we cannot dig into the LLMs to explore the reason for the effectiveness of using code-style instructions.
Furthermore, while we have investigated various designs for code-style instructions, there is still a need for further exploration of better prompt design.
Besides, querying the GPT-series models can lead to economic expenses and cause environmental pollution.

\section{Acknowledgements}
The authors wish to thank the anonymous reviewers for their helpful comments. This work was partially funded by National Natural Science Foundation of China (No.62206057,61976056,62076069), Shanghai Rising-Star Program (23QA1400200), Natural Science Foundation of Shanghai (23ZR1403500), Program of Shanghai Academic Research Leader under grant 22XD1401100, CCF-Baidu Open Fund, and CCF-Baichuan Fund.

\nocite{*}
\section{Bibliographical References}\label{sec:reference}

\bibliographystyle{lrec-coling2024-natbib}
\bibliography{lrec-coling2024-example}

\section{Language Resource References}
\label{lr:ref}
\bibliographystylelanguageresource{lrec-coling2024-natbib}
\bibliographylanguageresource{languageresource}

\appendix

\section*{Appendices}
\section{Datasets}
\label{sec:Datasets}
The statistics of the datasets in our paper have been presented in Table \ref{table: dataset}.
\subsection{AdvGLUE}
AdvGLUE \citeplanguageresource{wang2022advglue} is a multi-task benchmark to evaluate modern language models. The benchmark contains SST-2, QQP, MNLI, QNLI and RTE. It incorporates diverse forms of attacks at the word level, sentence level and also contains human-written samples.
SST-2 \citeplanguageresource{socher-etal-2013-sst2} consists of movie reviews with human-annotated sentiments. The task is to predict the sentiment of given sentences.
QQP is a collection of question pairs from the community question-answering website Quora. The task is to determine whether the given two questions are semantics equivalent.
MNLI \citeplanguageresource{williams-etal-2018-mnli} consists of a set of sentence pairs accompanied by annotations indicating textual entailment. The objective of this task is to determine whether a given premise sentence implies the hypothesis (\textit{entailment}), contradicts it (\textit{contradiction}), or has no clear relationship with it (\textit{neutral}).
QNLI \citeplanguageresource{rajpurkar-etal-2016-qnli} is a question-answering dataset consisting of question-paragraph pairs. The goal is to determine whether the paragraph contains the answer to the question.
RTE datasets come from a series of annual textual entailment challenges. The goal is to judge the relationships between two sentences, which include \textit{entailment} and \textit{not entailment}

\begin{table*}[htp]
    \centering
    \begin{tabular}{|c|c|c|c|}
    \hline
    Dataset & Task &Sample & Class \\
    \hline
    SST2 & sentiment classification & $148$ & $2$ \\
    \hline
    QQP & quora  question pairs & $78$ & $2$ \\
    \hline
    MNLI &multi-genre natural language inference & $121$ & $3$ \\
    \hline
    QNLI &question-answering NLI & $148$ & $2$ \\
    \hline
    RTE &textual entailment recognition & $81$ & $2$ \\
    \hline
    Restaurant-T & aspect-based sentiment analysis & $900$ & $3$ \\
    \hline
    \end{tabular}
    \caption{Statistics of test sets in this paper. For the Restaurant-T dataset, we randomly select 300 samples from each transformation (RevNon, RevTgt and AddDiff) and lead to a total of 900 samples.}
    \label{table: dataset}
\end{table*}

\subsection{Restaurant}
The Restaurant dataset is an Aspect-based sentiment analysis sourced from SemEvall 2014 dataset \citeplanguageresource{pontiki-etal-2014-semeval} and in this work, we use its adversarial version from \citeplanguageresource{xing2020tasty}. The adversarial transformation contains three parts: RevTgt, RevNon and AddDiff. RevTgt is to generate sentences that reverse the original sentiment of the target aspect.
RevNon aims to perturb the sentiments of the non-target aspects. Specifically, for all the non-target aspects with the same sentiment as the target aspects, we reverse their sentiments.
AddDiff further investigates if adding more non-target aspects can confuse the model. We add extra aspects that possess sentiments opposite to the target aspect.
In this work, we random select 300 samples from each transformation to conduct our experiments.

\begin{table*}
\centering
\resizebox{\linewidth}{!}{
\begin{tabular}{l|p{12cm}}
    \hline
    Dataset & Prompt \\
    \hline
    SST-2 & {Please classify the following sentence into either positive or negative. Answer me with "positive" or "negative", just one word.} \\
    \hline
    QQP & {Are the following two questions equivalent or not? Answer me with "equivalent" or "not\_equivalent".} \\
    \hline
    MNLI & {Are the following two sentences entailment, neutral or contradiction? Answer me with "entailment", "neutral" or "contradiction".} \\
    \hline
    QNLI & {Are the following question and sentence entailment or not\_entailment? Answer me with "entailment" or "not\_entailment".} \\
    \hline
    RTE & {Are the following two sentences entailment or not\_entailment? Answer me with "entailment" or "not\_entailment".} \\
    \hline
    Restaurant-T & {What is the sentiment towards '{sentence}' in terms of '{aspect word}'? Are they viewed positively, negatively, or neutrally?} \\
    \hline
\end{tabular}
}
\caption{natural language prompts}
\label{table: nl prompts}
\end{table*}
\section{Prompt Design}
\label{Appendix: Prompt}
% All the prompt design can be found in this section.
\subsection{Natural Language Prompts}
The natural language prompts of AdvGLUE are the same as \cite{wang2023robustness}. 
We present all the natural language prompts in Table \ref{table: nl prompts}.

\definecolor{dkgreen}{rgb}{0,0.6,0}
\definecolor{gray}{rgb}{0.5,0.5,0.5}
\definecolor{mauve}{rgb}{0.58,0,0.82}

\lstset{frame=tb,
  language=Python,
  aboveskip=3mm,
  belowskip=3mm,
  showstringspaces=false,
  columns=flexible,
  basicstyle={\small\ttfamily},
  numbers=none,
  numberstyle=\tiny\color{gray},
  keywordstyle=\color{blue},
  commentstyle=\color{dkgreen},
  stringstyle=\color{mauve},
  breaklines=true,
  breakatwhitespace=true,
  tabsize=3
}
\subsection{Code-style Prompts}
All prompts for the tasks in our paper will be presented in this section.

Prompts for SST-2:
\begin{lstlisting}
class Sentiment_Classification:
    """
    Base class for judging whether the sentiment of the given sentence is "positive" or "negative".
    
    Parameters
    ----------
    input_text : str
        The input sentence.

    """
    def __init__(self, input_text):
        self.input_text = input_text

    def sentiment_classification(self):
        polarity = self.input_text.sentiment.polarity

        if polarity > 0:
            print('positive')
        elif polarity < 0:
            print('negative')
\end{lstlisting}

Prompts for QQP:
\begin{lstlisting}
class Semantics_Consistent_Judgement:
    """
    Base class for judging whether the semantics of the two sentences are consistent.
    
    Parameters
    ----------
    input_text1 : str
        The first input sentence.
    input_text2 : str
        The second input sentence.
    """
    def __init__(self, input_text1, input_text2):
        self.input_text1 = input_text1
        self.input_text2 = input_text2
    
    def semantics_similarity(self):

        similarity = cosine_similarity(self.input_text1, self.input_text2)
    
        if similarity > 0:
            print("equivalent")
        elif similarity < 0:
            print("not_equivalent")
\end{lstlisting}

Prompts for MNLI:
\begin{lstlisting}
class Entailment_Judgement:
    """
    Base class for judging whether the premise and the hypothesis are "entailment", "neutral" or "contradiction".
    
    Parameters
    ----------
    premise : str
        The input premise.
    hypothesis : str
        The input hypothesis.
    """
    def __init__(self, premise: str, hypothesis: str):
        self.premise = premise
        self.hypothesis = hypothesis
    
    def determine_relationship(self):
        if is_entailment(self.premise, self.hypothesis):
            print("entailment")
        elif is_contradiction(self.premise, self.hypothesis):
            print("contradiction")
        else:
            print("neutral")
\end{lstlisting}

Prompts for QNLI:
\begin{lstlisting}
class Answer_Verification:
    """
    Given a question, determines whether the provided text contains the correct answer to the question.
    The relationship consists of "entailment" and "not entailment".
    
    Parameters
    ----------
    question : str
        The input question.
    text : str
        The input text.
    """
    def __init__(self, question, text):
        self.question = question
        self.text = text

    def determine_relationship(self):
        if is_entailment(self.question, self.text):
            print("entailment")
        else:
            print("not_entailment")
\end{lstlisting}

Prompts for RTE:
\begin{lstlisting}
class Entailment_Judgement:
    """
    Base class for judging whether the two sentences are "entailment" or "not_entailment".
    
    Parameters
    ----------
    sentence1 : str
        The first input sentence.
    sentence2 : str
        The second input sentence.
    """
    def __init__(self, premise: str, hypothesis: str, relationship: str):
        self.sentence1 = sentence1
        self.sentence2 = sentence2
    
    def determine_relationship(self):
        if is_entailment(self.sentence1, self.sentence2):
            print("entailment")
        else:
            print("not_entailment")
\end{lstlisting}

Prompts for Restaurant-T:
\begin{lstlisting}
class Aspect_Based_Sentiment_Analysis:
    """
    Base class for aspect-based sentiment analysis task.
    
    Parameters
    ----------
    aspect : str
        The target aspect term of the given sentence.
    sentence : str
        The input text that contains the aspect.
    """
    def __init__(self, sentence: str, aspect: str):
        self.sentence = sentence
        self.aspect = aspect
    def sentiment_classification(self):
    if is_positive(self.sentence, self.aspect):
        print("positive")
    elif is_negative(self.sentence, self.aspect):
        print("negative")
    
\end{lstlisting}

\subsection{Other Code-style prompts Design}
\label{Appecdix B.3:Different code-style instructions}
In Section \ref{5.3}, we design two other different code-style prompts: \textit{class init} and \textit{func exec}.
Specifically, the "class init" prompt provides LLMs with incomplete code with partial parameter input of the class as triggers to prompt LLMs to complete the code. The "func exec" converts the class definition into a function definition. The detailed results of these two designs can be found in Table 2. Besides, following your advice, we will also add other ablation experiments with regard to the prompt design in the subsequent version of our paper, such as the influence of certain parts of the prompt (Class name, annotation etc.)
We use the QNLI task as an example to present the different prompts.

Prompts for \textit{class init}:
\begin{lstlisting}
class Answer_Verification:
    """
    Given a question, determines whether the provided text contains the correct answer to the question.
    The relationship consists of "entailment" and "not entailment".
    
    Parameters
    ----------
    question : str
        The input question.
    text : str
        The input text.
    """
    def __init__(self, question, text, relationship):
        self.question = question
        self.text = text
        self.relationship = relationship
\end{lstlisting}

Prompts for \textit{func exec}:
\begin{lstlisting}
def Answer_Verification(question: str, text: str):
    """
    Given a question, determines whether the provided text contains the correct answer to the question.
    The relationship consists of "entailment" and "not entailment".

    Args:
        question (str): The input question.
        text (str): The input text.

    Returns:
        str: "entailment", or "not entailment".
    """

    if is_entailment(question, text):
        print("entailment")
    else:
        print("not_entailment")
\end{lstlisting}

\section{Detailed Experiments Results}
\label{Appendix B: Detailed Experiments Results}
The detailed experiments of accuracy can be found in this section.
Table \ref{table: Origin results of text-davinci-003} and Table \ref{table: Origin results of gpt-3.5-turbo} present the accuracy of datasets before (Original) and after adversarial transformations (Adversarial) with \texttt{text-davinci-003}. 
Moreover, Table \ref{table: Origin results of gpt-3.5-turbo} and Table \ref{table: Adversarial results of gpt-3.5-turbo} present the results of gpt-3.5-turbo.
All the results are under few-shot settings.
From Tabel \ref{table: Origin results of text-davinci-003} and Table \ref{table: Adversarial results of text-davinci-003}, we can conclude that using code-style prompts only acquire little improvement on the original datasets. Specifically, using code-style instructions outperforms using natural language instructions by 5.13 and 5.68 points in accuracy with \texttt{text-davinci-003} and \texttt{gpt-3.5-turbo}, respectively.
However, when employing adversarial samples in the instructions, using code-style instructions acquires a relatively larger improvement. 
Specifically, we get 3.02 and 5.68 points with \texttt{text-davinci-003} and \texttt{gpt-3.5-turbo}, respectively, showcasing the advantages of code-style instructions in resisting adversarial attacks.

\begin{table*}[htp]
\centering
\resizebox{\linewidth}{!}{
\begin{tabular}{c|ccccc|ccc|c}
\hline
  \multirow{4}{*}{\textbf{Method}} & \multicolumn{8}{c|}{\multirow{2}{*}{\textbf{Dataset}}} & \multirow{4}{*}{\textbf{AVG}}\\
& \multicolumn{8}{c|}{} &\\
\cline{2-9}
& \multicolumn{5}{c|}{\textbf{AdvGLUE}} & \multicolumn{3}{c|}{\textbf{Restaurant-T}} &\\

& \textbf{SST-2} & \textbf{QQP} & \textbf{MNLI} & \textbf{QNLI} & \textbf{RTE} & \textbf{RevTgt} & \textbf{RevNon} & \textbf{AddDiff} &\\
\hline

NL & $96.18$ & $82.05$ & $79.33$ & $79.72$ & $92.59$ & $91.66$ & $92.67$ & $80$ & $86.78$\\
Code & $98.47$ & $83.33$ & $84.2$ & $77.7$ & $92.59$ & $91$& $92$& $78.67$ & $87.25$\\
\hline
    
\end{tabular}
}
\caption{Original datasets results with text-davinci-003}
\label{table: Origin results of text-davinci-003}
\end{table*}

\begin{table*}
\centering
\resizebox{\linewidth}{!}{
\begin{tabular}{c|ccccc|ccc|c}
\hline
  \multirow{4}{*}{\textbf{Method}} & \multicolumn{8}{c|}{\multirow{2}{*}{\textbf{Dataset}}} & \multirow{4}{*}{\textbf{AVG}}\\
& \multicolumn{8}{c|}{} &\\
\cline{2-9}
& \multicolumn{5}{c|}{\textbf{AdvGLUE}} & \multicolumn{3}{c|}{\textbf{Restaurant-T}} &\\

& \textbf{SST-2} & \textbf{QQP} & \textbf{MNLI} & \textbf{QNLI} & \textbf{RTE} & \textbf{RevTgt} & \textbf{RevNon} & \textbf{AddDiff} &\\
\hline

NL & $70.94$ & $67.95$ & $61.15$ & $63.51$ & $82.72$ & $67$ & $\textbf{86}$ & $76$ & $72.07$\\
CoT & $74.32$ & $71.79$ & $70.25$ & $\textbf{67.57}$ & $81.48$ & $70.66$ & $82$& $69.33$ &  $73.43$\\
\textbf{Code} & $75$ & $73.07$ & $68.59$ & $66.21$ & $\textbf{87.65}$ & $70.67$ & $\textbf{88}$ & $77.33 $& $75.82$\\
\textbf{Code+adv} & $\textbf{78.38}$ & $\textbf{75.64}$ & $\textbf{72.72}$ & $64.86$ & $\textbf{87.65}$ & $\textbf{72.66}$ & $87.33$ & $\textbf{78.33}$& $\textbf{77.20}$\\

\hline
    
\end{tabular}
}
\caption{Adversarial datasets results of text-davinci-003}
\label{table: Adversarial results of text-davinci-003}
\end{table*}

\begin{table*}
\centering
\resizebox{\linewidth}{!}{
\begin{tabular}{c|ccccc|ccc|c}
\hline
  \multirow{4}{*}{\textbf{Method}} & \multicolumn{8}{c|}{\multirow{2}{*}{\textbf{Dataset}}} & \multirow{4}{*}{\textbf{AVG}}\\
& \multicolumn{8}{c|}{} &\\
\cline{2-9}
& \multicolumn{5}{c|}{\textbf{AdvGLUE}} & \multicolumn{3}{c|}{\textbf{Restaurant-T}} &\\

& \textbf{SST-2} & \textbf{QQP} & \textbf{MNLI} & \textbf{QNLI} & \textbf{RTE} & \textbf{RevTgt} & \textbf{RevNon} & \textbf{AddDiff} &\\
\hline

NL & $99.23$ & $83.33$ & $76.03$ & $75.67$ & $83.95$ & $91.33$ & $92$ & $83.33$ & $85.61$\\
Code & $98.47$ & $83.33$ & $80.16$ & $75.67$ & $86.42$ & $91.33$ & $92$ & $81.67$ & $86.13$\\
\hline
    
\end{tabular}
}
\caption{Original datasets results of gpt-3.5-turbo}
\label{table: Origin results of gpt-3.5-turbo}
\end{table*}

\begin{table*}
\centering
\resizebox{\linewidth}{!}{
\begin{tabular}{c|ccccc|ccc|c}
\hline
  \multirow{4}{*}{\textbf{Method}} & \multicolumn{8}{c|}{\multirow{2}{*}{\textbf{Dataset}}} & \multirow{4}{*}{\textbf{AVG}}\\
& \multicolumn{8}{c|}{} &\\
\cline{2-9}
& \multicolumn{5}{c|}{\textbf{AdvGLUE}} & \multicolumn{3}{c|}{\textbf{Restaurant-T}} &\\

& \textbf{SST-2} & \textbf{QQP} & \textbf{MNLI} & \textbf{QNLI} & \textbf{RTE} & \textbf{RevTgt} & \textbf{RevNon} & \textbf{AddDiff} &\\
\hline
NL & $79.05$ & $76.92$ & $60.33$ & $62.83$ & $67.9$ & $67.3$ & $77.3$ & $72$ & $70.45$\\
CoT & $78.37$ & $73.07$ & $\textbf{74.38}$ & $64.18$ & $69.13$ & $62$ & $79$ & $69$ & $71.14$\\
\textbf{Code} & $83.1$ & $74.35$ & $69.42$ & $73.64$ & $\textbf{74.07}$ & $71.33$ & $79.6$ & $\textbf{72.33}$ & $74.73$\\
\textbf{Code+adv} & $\textbf{83.78}$ & $\textbf{82.05}$ & $72.73$ & $\textbf{74.32}$ & $72.83$ & $\textbf{72.33}$ & $\textbf{80}$ & $71$ & $\textbf{76.13}$\\

\hline
    
\end{tabular}
}
\caption{Adversarial datasets results of gpt-3.5-turbo}
\label{table: Adversarial results of gpt-3.5-turbo}
\end{table*}

\end{document}